\documentclass[letterpaper]{article} 
\usepackage{aaai2027}  
\usepackage[hyphens]{url}  
\usepackage{graphicx} 
\def\UrlFont{\rm}  
\usepackage{natbib}  
\usepackage{caption} 
\usepackage{algorithm}
\usepackage{algorithmic}
\usepackage{array}
\usepackage{booktabs}
\usepackage{tabularx}
\usepackage{amsmath}
\usepackage{multirow}
\usepackage{amssymb}
\usepackage{xcolor}
\usepackage{newfloat}
\usepackage{listings}
\DeclareCaptionStyle{ruled}{labelfont=normalfont,labelsep=colon,strut=off} 
\floatstyle{ruled}
\newfloat{listing}{tb}{lst}{}
\floatname{listing}{Listing}

\usepackage{booktabs}

\title{From Transcription to Reformulation: Defining and Benchmarking Grounded Generative Speech Recognition}
\author{
    Qing Wang\textsuperscript{\rm 1}\equalcontrib,
    Qihan Hu\textsuperscript{\rm 1}\equalcontrib,
    Hang Lv\textsuperscript{\rm 1},
    Yan Yang\textsuperscript{\rm 1},
    Boqing Liu\textsuperscript{\rm 1},
    Xuechao Zhu\textsuperscript{\rm 1},
    Zehan Li\textsuperscript{\rm 1},
    Hongjie Chen\textsuperscript{\rm 1},
    Jian Kang\textsuperscript{\rm 1},
    Jie Li\textsuperscript{\rm 1},
    Yongxiang Li\textsuperscript{\rm 1}\corresponding,
    Lei Xie\textsuperscript{\rm 2}\corresponding
}
\affiliations{
    \textsuperscript{\rm 1}Xingchen AGI Lab, China Telecom Artificial Intelligence Technology Co. Ltd, China\\
    \textsuperscript{\rm 2}Audio, Speech and Language Processing Group (ASLP@NPU), School of Computer Science, \\Northwestern Polytechnical University, Xi’an, China

}

\begin{document}

\maketitle

\begin{abstract}

Speech language models have expanded automatic speech recognition (ASR) from transcription to general speech‑to‑text generation. Yet existing benchmarks still assume a single ground‑truth transcript per utterance—an assumption that struggles when models produce readable, formatted, or structured outputs.
In this work, we introduce Grounded \textbf{Generative Speech Recognition (GSR)}, reformulating speech recognition as generating semantically grounded text instead of a single transcript. We define a \textit{grounded reformulation space} that characterizes all valid textual realizations supported by the input speech, and establish a four-level hierarchy of reformulation capabilities, including literal transcription, canonical reformulation, semantic reformulation and speech-grounded structuring.
Based on this formulation, we present \textbf{GSR-Bench}, the first comprehensive benchmark designed to evaluate generative speech recognition systems beyond exact transcript matching. GSR-Bench covers diverse scenarios including refined meeting transcription, cross-dialect translation, context-aware speech understanding, and robustness under silent or noisy acoustic conditions. To address the limitations of word-level evaluation metrics, we develop a multi-dimensional evaluation framework combining speech grounding assessment and LLM-based semantic judgment, whose reliability is validated through human preference studies.
Extensive experiments across representative architectures, ranging from conventional ASR systems to end-to-end Speech-LLMs, reveal fundamental trade-offs between transcription fidelity and generative reformulation ability. Our analysis further identifies critical failure modes, including acoustic hallucination, unsupported generation, and instruction drift, providing insights into the design and evaluation of future grounded generative speech systems.

\end{abstract}


\section{Introduction}
Automatic speech recognition (ASR)~\cite{Graves2012,Chan2016,gulati2020conformer,yao2023zipformer,kim2022squeezeformer,kim2023branchformer} has traditionally been formulated as an speech-to-text transcription problem, where the primary objective is to recover the exact linguistic content contained in speech. Over the past decades, advances in neural architectures, learning criteria, self-supervised speech representation learning~\cite{baevski2020wav2vec,baevski2020vq,hsu2021hubert,chen2022wavlm}, ultra-large-scale multi-task pre-training, and convenient and systematic training frameworks have substantially improved transcription accuracy, leading to near-human performance on many benchmark datasets.

However, the emergence of Speech Language Models (SLM)~\cite{rubenstein2023audiopalm,zhang2023speechgpt,chu2023qwen,barrault2023seamlessm4t}are fundamentally changing the role of speech recognition. Instead of merely producing literal transcriptions, modern speech models are increasingly capable of generating readable, formatted, and task-oriented textual outputs, including normalized transcripts, translated content, meeting summaries, and structured information. This evolution suggests that speech recognition is transitioning from a transcription problem toward a more general speech-grounded generation paradigm.

Despite this progress, current definitions and evaluation methodologies for speech recognition remain largely inherited from the transcription paradigm. Conventional ASR assumes that each speech utterance corresponds to a unique ground-truth transcript, and therefore evaluates systems using exact matching metrics such as word error rate (WER) and character error rate (CER).

While effective for measuring transcription accuracy, these metrics become insufficient when models intentionally perform legitimate textual reformulations. For example, removing disfluencies ~\cite{Kim2021}, restoring punctuation, normalizing spoken expressions ~\cite{Shi2024}, or generating structured summaries ~\cite{LaQuatra2025} may produce outputs that differ from the reference transcript while preserving the intended meaning. Therefore, exact transcript matching can incorrectly penalize desirable generations while failing to measure whether generated content is actually grounded in the input speech.

To address this limitation, we argue that speech recognition should be reconsidered from the perspective of \textit{grounded reformulation}. In this work, we introduce \textbf{Generative Speech Recognition (GSR)}, a new paradigm that defines speech recognition as the generation of appropriate textual realizations grounded in the input speech rather than the recovery of a single transcript.

Unlike conventional ASR, where the output space is restricted to the literal transcription, GSR considers a set of valid semantic-preserving reformulations. These reformulations may vary in abstraction level and presentation style according to different task objectives, while maintaining faithful grounding to the original speech. Based on this perspective, we establish a hierarchical taxonomy consisting of four levels: literal transcription, canonicalized transcription, semantic reformulation, and speech-grounded structuring.

Building upon this formulation, we introduce \textbf{GSR-Bench}, a comprehensive benchmark for evaluating grounded generative speech systems. Unlike existing ASR benchmarks~\cite{Chen2024AIRBench,arora2025talking,chen2026voicebench} that primarily measure transcription accuracy, GSR-Bench evaluates models across multiple dimensions of generative capability, including reformulation quality, speech grounding, instruction following, and robustness under challenging acoustic conditions.

The benchmark contains diverse evaluation suites covering refined meeting transcription, cross-dialect translation, context-aware generation, and boundary cases involving silence and noise. Furthermore, we develop a semantic evaluation pipeline based on LLM-as-a-judge, which goes beyond exact matching metrics by assessing whether generated outputs remain faithful to the underlying speech information.

We benchmark representative systems ranging from conventional ASR pipelines to modern Speech-LLMs and reveal several important observations. First, transcription-oriented models achieve strong literal accuracy but struggle with higher-level reformulation tasks. Second, generative models demonstrate improved readability and task adaptability but introduce new risks, including acoustic hallucination, unsupported generation, and instruction drift. These findings highlight the necessity of evaluating speech systems under the broader perspective of grounded generation.

Our contributions are summarized as follows:
\begin{itemize}
    \item We introduce and formally define grounded \textbf{Generative Speech Recognition (GSR)}, re-framing speech recognition from transcription
    to grounded textual reformulation.
    \item We establish a four-level \textbf{Reformulation Hierarchy} that provides a unified view of emerging speech generation capabilities.
    \item We propose \textbf{GSR-Bench}, a comprehensive benchmark and evaluation framework for systematically measuring generative speech recognition systems.
    \item We conduct extensive analysis across representative ASR and SLM architectures, revealing the trade-offs between transcription fidelity, reformulation ability, and grounding reliability.
\end{itemize}

\section{Generative Speech Recognition}

\begin{figure*}[t]
\centering
\includegraphics[width=0.85\textwidth]{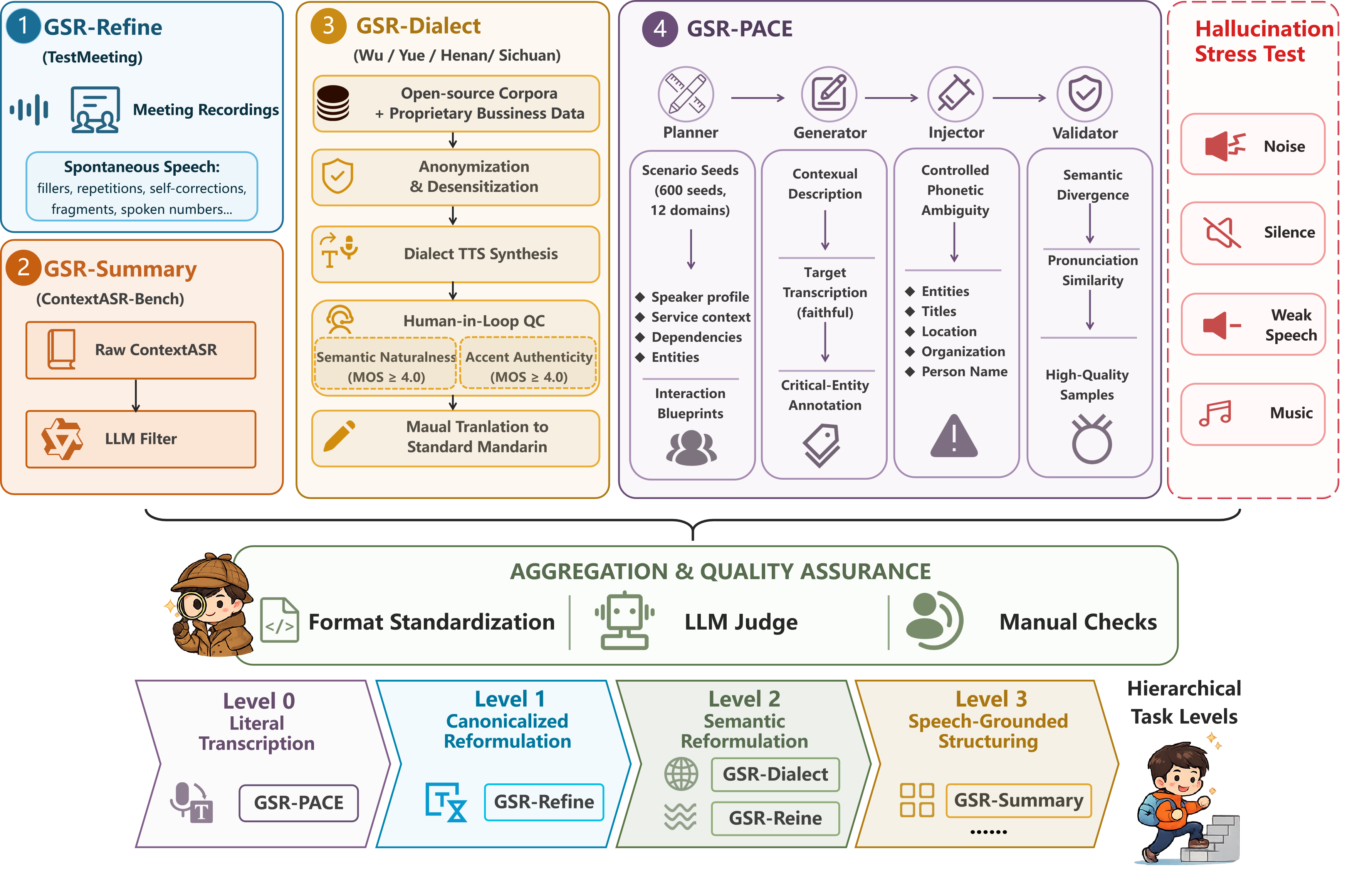} 
\caption{Overview of GSR-Bench: Data Construction, Quality Assurance, and Hierarchical Task Organization.}
\label{fig1}
\end{figure*}

\subsection{From Transcription to Reformulation} 

Conventional automatic speech recognition (ASR) systems formulate speech recognition as a sequence transduction problem, where the primary objective is to recover a textual sequence that maximally matches the spoken content. Given speech signal $\mathbf{S}$, conventional ASR aims to generate a transcript $\mathbf{T}^{*}$ by optimizing the likelihood:
\begin{equation}
\mbox{\small$\displaystyle
\mathbf{T}^{*}
=
\arg\max_{\mathbf{Y}}
P(\mathbf{Y}|\mathbf{S}),
$}
\end{equation}
where the generated output is expected to literally transcribe the speech under a strict 'what-you-hear-is-what-you-get' principle. As a result, standard evaluation protocols (e.g., WER and CER) strictly rely on human-annotated gold ground-truth transcripts, assuming that a single reference text represents the sole correct output.

However, recent speech language models have demonstrated the ability to generate text beyond literal transcription. Instead of simply recovering spoken words, these models can perform a wide range of reformulations, including punctuation restoration, inverse text normalization (ITN), disfluency removal, paraphrasing, translation, and structured information extraction~\cite{Du2023,Teleki2025}. Such outputs may differ significantly from the literal speech transcription, while remaining faithful to the information conveyed by the speaker.

We argue that this capability represents a fundamental paradigm shift from \textit{transcription} to \textit{grounded reformulation}. Therefore, we define GSR as a speech-to-text paradigm that models recognition as the generation of task-specific textual representations grounded by the acoustic input, rather than the recovery of a single literal transcription.

Unlike conventional ASR, where any non-literal output is considered an error, GSR allows flexible text formulations that capture the intended meaning.
The key point is \textbf{speech grounding}: every generated output must remain supported by the input speech and any explicitly provided contextual information.

\subsection{Grounded Reformulation Space}

To formally characterize the output flexibility of the GSR, we first define the grounded reformulation space. Given a speech signal $\mathbf{S}$, we define the grounded reformulation space as
\vspace{-0.7em}
\begin{equation}
\mbox{\small$\displaystyle
\mathcal{R}(\mathbf{S})
=
\left\{
\mathbf{Y}
\mid
G(\mathbf{Y},\mathbf{S})\geq\gamma
\right\},
\label{eq:reformulation_space}
$}
\end{equation}
where $\mathbf{Y}$ denotes a generated textual representation, $G(\cdot,\cdot)$ represents a grounding function measuring whether the generated output is based on the source input speech and $\gamma$ denotes the minimum grounding threshold.

Therefore, when the grounding constraint is violated, represented by $G(\mathbf{Y},\mathbf{S})<\gamma$, the GSR system tends to yield unverified outputs driven by its linguistic priors, which naturally describes the phenomenon of hallucination.


In practical scenarios, GSR systems are often controlled by task instructions or contextual information. Let $\mathbf{P}$ denote a generation instruction and $\mathbf{C}$ denote the auxiliary context. To capture the multi-modal and situational nature of speech applications, we formulate $\mathbf{C}$ as a structured set of contextual priors:
\vspace{-0.75em}
\begin{equation}
\mbox{\small$\displaystyle
\mathbf{C} = \{\mathbf{C}_{\mathrm{conversational}}, \mathbf{C}_{\mathrm{environmental}}, \mathbf{C}_{\mathrm{meta}}\}
$}
\end{equation}
where $\mathbf{C}_{\mathrm{conversational}}$ captures prior dialogue history or preceding text, $\mathbf{C}_{\mathrm{environmental}}$ represents acoustic background profiles or prior audio, and $\mathbf{C}_{\mathrm{meta}}$ represents metadata such as speaker profiles, scenario types, or entity hotwords. These conditions define a task-specific subspace:
\begin{equation}
\mbox{\small$\displaystyle
\mathcal{R}(\mathbf{S},\mathbf{P},\mathbf{C})
\subseteq
\mathcal{R}(\mathbf{S}),
\label{eq:conditional_space}
$}
\end{equation}
where $\mathbf{P}$ determines the desired reformulation objective and $\mathbf{C}$ provides additional evidence for generation.

Importantly, contextual information does not itself increase the reformulation level. Context only changes the available evidence for generation, while the degree of permissible reformulation is determined by the task objective.

\subsection{Reformulation Hierarchy}

Although GSR outputs must strictly satisfy the grounding constraint, different downstream tasks permit varying degrees of reformulation from the spoken surface form. 
To formalize this progression, we extend the \textbf{Grounded Reformulation Space} by introducing a hierarchical organization based on Human Comprehensibility ($H$), structured according to the degree of semantic and structural transformation.

Specifically, we evaluate Human Comprehensibility $H(\mathbf{Y})$ through a multi-dimensional composite metric that integrates five core aspects: Faithfulness ($F$), Coverage ($C$), Key Information Preservation ($K$), Conciseness ($N$), and Readability ($R$). Formally, $H(\mathbf{Y})$ is defined as:
\mbox{\small$\displaystyle
$$H(\mathbf{Y}) = \sum_{i=1}^{5} w_i M_i(\mathbf{Y})$$
$}
where $\mathbf{M} = [F, C, K, N, R]^T$, and $w_i \ge 0$ are the corresponding weights satisfying $\sum_{i=1}^{5} w_i = 1$. Notably, this five-dimensional framework is extensible, with plans to expand its metrics in future iterations.

From the evaluation perspective, based on above metrics, the \textbf{Reformulation Hierarchy} is defined by parameterizing the target comprehensibility threshold $h_l$ for each level $l \in \{0, 1, 2, 3\}$. Given an input speech $\mathbf{S}$, the admissible output space at level $l$, denoted as $\mathcal{R}_l(\mathbf{S})$, is formulated as:
\begin{equation}
\mbox{\small$\displaystyle
$$\mathcal{R}_l(\mathbf{S}) = \left\{ \mathbf{Y} \;\Big\vert{}\; G(\mathbf{Y}, \mathbf{S}) \ge \gamma \land H(\mathbf{Y}) \ge h_l \right\}$$
$}
\end{equation}
where $h_l$ represents the minimum comprehensibility requirement for level $l$. 
Mathematically, we define the hierarchy of these spaces as a monotonic expansion of generative degrees of freedom:
\vspace{-0.25em}
\begin{equation}
\mbox{\small$\displaystyle
\mathcal{R}_3(\mathbf{S})
\subset
\mathcal{R}_2(\mathbf{S})
\subset
\mathcal{R}_1(\mathbf{S})
\subset
\mathcal{R}_0(\mathbf{S})
\subseteq
\mathcal{R}(\mathbf{S}),
\label{eq:hierarchy}
$}
\end{equation}
\vspace{-0.25em}
where higher levels permit increasingly flexible reformulations and more fruitful information for understanding. Note that this hierarchy represents output flexibility rather than task difficulty.

Corresponding the evaluation perspective, this following section provides a detailed breakdown of each level within the \textbf{Reformulation Hierarchy} from the standpoint of practical implementation.

\paragraph{Level 0: Literal Transcription.}
Level 0 corresponds to the conventional ASR setting, where no reformulation is allowed. The objective is to recover the spoken content with maximum lexical fidelity, preserving the original vocabulary, word order, repetitions, hesitations, and spontaneous speech phenomena whenever they are acoustically observable. Formally, the output space is restricted to:
\begin{equation}
\mbox{\small$\displaystyle
\mathcal{R}_0(\mathbf{S})
=
\{
\mathbf{Y}
|
\mathbf{Y}
\approx
\mathbf{T}
\},
$}
\end{equation}
where $\mathbf{T}$ denotes the literal transcript of the speech. This level serves as the recognition baseline for both ASR-specialized systems and GSR models.

\paragraph{Level 1: Canonical Reformulation.}
Level 1 permits surface-form reformulations while preserving the complete semantic content of the utterance. The model may modify writing conventions, including punctuation restoration, sentence segmentation, capitalization, and inverse text normalization (ITN). The reformulation can be represented as:
\begin{equation}
\mbox{\small$\displaystyle
\mathcal{R}_1(\mathbf{S})
=
\{
f(\mathbf{T})
|
f\in\mathcal{F}_{surface}
\},
$}
\end{equation}
where $\mathcal{F}_{surface}$ denotes a set of surface-level rewriting operations. At this level, entities, numerical values, temporal expressions, negations, and logical relations must remain unchanged. The difference from conventional ASR post-processing lies in controllability: GSR performs canonical reformulation as a generation objective rather than applying fixed normalization rules after recognition.

\paragraph{Level 2: Semantic Reformulation.}
Level 2 further extends the reformulation capability from surface modification to meaning-preserving reformulation. The generated output may differ from the original transcript in wording, sentence structure, or language, provided that the semantic content remains unchanged. The admissible space is defined as:
\begin{equation}
\mbox{\small$\displaystyle
\mathcal{R}_2(\mathbf{S})
=
\{
f(\mathbf{T})
|
f\in\mathcal{F}_{sem},
\operatorname{Sem}(f(\mathbf{T}))
=
\operatorname{Sem}(\mathbf{T})
\}
$}
\end{equation}
where $\mathcal{F}_{sem}$ includes operations such as disfluency removal, paraphrasing, dialect normalization, and speech translation. Compared with Level 1, this level shifts the preservation target from lexical form to semantic content. The model is encouraged to produce clearer and more task-appropriate language while preserving all information explicitly conveyed by the speaker.

\paragraph{Level 3: Speech-grounded Structuring.}
Level 3 represents the highest degree of reformulation, where speech is transformed into structured semantic representations rather than transcript-like text. The output space is defined as:
\begin{equation}
\mbox{\small$\displaystyle
\mathcal{R}_3(\mathbf{S})
=
\{
f(\mathbf{T})
|
f\in\mathcal{F}_{structure}
\},
$}
\end{equation}
where $\mathcal{F}_{structure}$ contains representation-level reformulations, including information extraction, intent representation, meeting summarization, speaker-attributed transcripts, and structured JSON generation. Unlike lower levels, Level 3 does not require sentence-level correspondence with the original speech. However, every generated fact, attribute, and relation must remain directly supported by the speech or explicitly authorized context. Open-ended dialogue generation and responses requiring external world knowledge are outside the scope of GSR.

\subsection{Probabilistic Formulation of GSR}

Based on the grounded reformulation space and reformulation hierarchy above, we formally define Generative Speech Recognition as follows: Given an speech input $\mathbf{S}$, an instruction prompt $\mathbf{P}$, and contextual information $\mathbf{C}$, a GSR system generates a textual representation:
\vspace{-0.25em}
\begin{equation}
\mbox{\small$\displaystyle
\mathbf{Y}^{*}
=
\arg\max_{\mathbf{Y}
\in
\mathcal{R}(\mathbf{S},\mathbf{P},\mathbf{C})}
P(
\mathbf{Y}
|
\mathbf{S},
\mathbf{P},
\mathbf{C}
).
\label{eq:gsr_definition}
$}
\end{equation}
\vspace{-0.25em}
For autoregressive generation, the probability can be factorized as:
\vspace{-0.75em}
\begin{equation}
\mbox{\small$\displaystyle
P(
\mathbf{Y}
|
\mathbf{S},
\mathbf{P},
\mathbf{C}
)
=
\prod_{u=1}^{U}
P(
y_u
|
y_{<u},
\mathbf{S},
\mathbf{P},
\mathbf{C}
),
$}
\end{equation}
\vspace{-0.25em}
where $y_u$ denotes the $u$-th generated token and $y_{<u}$ represents previously generated tokens.

The fundamental difference between GSR and conventional ASR lies in the optimization target. ASR seeks a literal transcription that minimizes lexical discrepancy, whereas GSR seeks an instruction-controlled reformulation within a grounded output space. Therefore, GSR can simultaneously optimize readability, structure, and task adaptability while preserving acoustic faithfulness.

Consequently, GSR evaluation should independently measure reformulation capability and grounding reliability, which motivates the design of GSR-Bench.

\section{GSR-Bench: A Benchmark for Grounded Generative Speech Recognition}

\subsection{Benchmark Overview}

\begin{figure*}[t]
\centering
\includegraphics[width=0.85\textwidth]{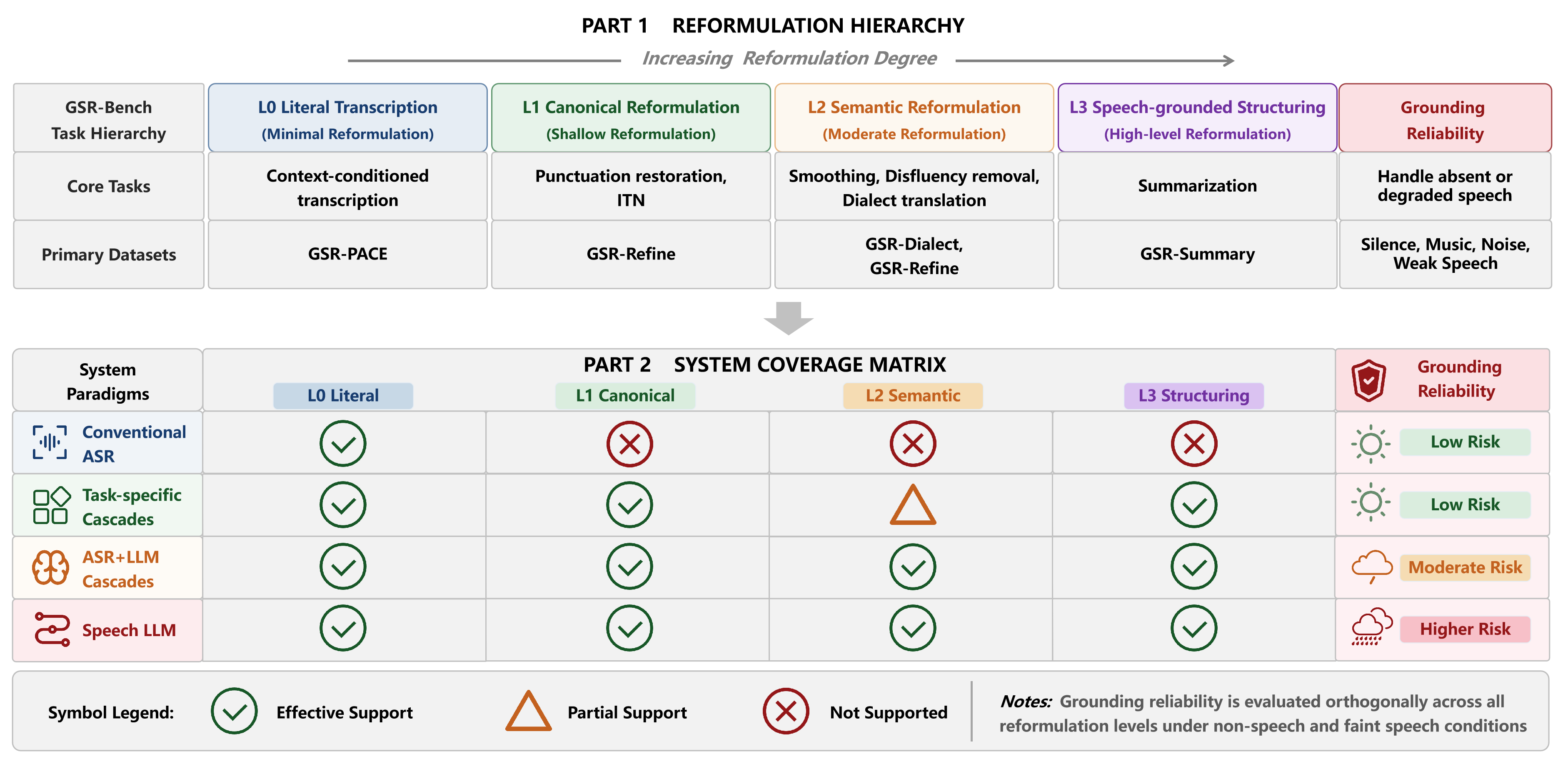} 
\caption{GSR-Bench Evaluation Framework: Reformulation Hierarchy, System Coverage, and Grounding Reliability.}
\label{fig2}
\end{figure*}

Existing ASR benchmarks rely on WER or CER under the assumption of a single literal reference, whereas GSR admits multiple valid speech-grounded reformulations and therefore requires joint evaluation of transcription fidelity, controlled reformulation, and grounding reliability.

To this end, we introduce \textbf{GSR-Bench}, a comprehensive benchmark designed to evaluate generative speech recognition systems from two orthogonal perspectives:

\begin{enumerate}
    \item \textbf{Reformulation Capability}: measuring how effectively a model transforms speech into increasingly flexible textual representations according to the proposed hierarchy, from literal transcription to structured semantic representations.

    \item \textbf{Grounding Reliability}: measuring whether generated outputs remain supported by acoustic evidence under challenging conditions, including non-speech audio, silence, and severely degraded speech.
\end{enumerate}

The following sections introduce the datasets and evaluation tasks used to instantiate the four reformulation levels, followed by a separate grounding-reliability stress test. Detailed dataset construction procedures and evaluation protocols are provided in the Appendix

\subsubsection{GSR-PACE:}

To instantiate \textbf{Level 0} of the proposed reformulation hierarchy, we introduce GSR-PACE (Phonetic Ambiguity with Contextual Evidence), a literal transcription dataset that tests whether models leverage auxiliary context to resolve phonetically confusable expressions. It explicitly constructs homophones, near‑homophones, pinyin‑similar variants, and commonly confused named entities around critical semantic anchors, while requiring the model to preserve the source speech’s literal lexical content. As formalized in Section 3, context context may aid disambiguation but does not authorize rewriting, omission, or semantic expansion.

Each sample pairs an ambiguous utterance with context that resolves the intended entity over phonetically similar alternatives. Focusing on entity‑rich domains (names, organizations, locations, titles, terms), the dataset evaluates contextual disambiguation while still penalizing models that override reliable acoustics with misleading context. We formulate the construction of GSR-PACE as a four-stage pipeline consisting of \textit{Planner}, \textit{Generator}, \textit{Injector}, and \textit{Validator}.

\textit{Planner: }The planner converts manually designed scenario seeds into realistic interaction blueprints. Each blueprint contains scenario descriptions, user profiles, contextual dependencies, and candidate entities. Unlike isolated keyword injection, this process ensures that every target entity appears naturally within a coherent real-world context.

\textit{Generator: }
Given each blueprint, the generator produces contextual descriptions, target utterances, and entity annotations. The generated utterance follows the literal transcription objective: all entities must correspond to the intended spoken content, while contextual information provides additional evidence for resolving ambiguities.

\textit{Injector: }
To simulate realistic recognition challenges, the injector introduces controlled acoustic ambiguities around semantic anchors. The injected confusions include homophones, near-homophones, pinyin-similar expressions, and commonly confused named entities.
The resulting samples evaluate whether models can appropriately leverage context rather than relying solely on acoustic similarity.

\textit{Validator: }
The validator filters generated samples according to four criteria: sufficient phonetic similarity between the target and confusing candidates; meaningful semantic difference; contextual resolvability; structural consistency.
Only samples satisfying all constraints are retained.

\subsubsection{GSR-Refine:}

To instantiate \textbf{Levels 1 and 2} of the proposed reformulation hierarchy, we introduce GSR-Refine, a multi-level benchmark constructed from the TestMeeting subset of WenetSpeech~\cite{zhang2022wenetspeech}. GSR-Refine evaluates the same spontaneous meeting recordings under two task instructions: Level 1 assesses canonical reformulation, whereas Level 2 further evaluates meaning-preserving semantic reformulation. By holding the acoustic input fixed, this design isolates the effect of increasing reformulation freedom from variations in speech content and recording conditions.

Compared with read or scripted speech, spontaneous meeting speech contains frequent fillers, repetitions, false starts, self-corrections, fragmented expressions, and spoken-form numerical expressions. These characteristics make TestMeeting suitable for evaluating whether models can transform natural speech into standardized and readable written text while faithfully preserving the information conveyed by the speaker.

At Level 1, models perform canonical reformulation, including punctuation restoration, sentence segmentation, and inverse text normalization. Although the output may differ from the literal transcript in written form, named entities, numerical values, temporal expressions, negations, and logical relations must remain unchanged. At Level 2, models are further allowed to remove semantically empty fillers, merge unintended repetitions, resolve explicit self-corrections, and reorganize fragmented expressions into fluent written text. The same acoustic inputs are used for both levels with different task instructions, enabling controlled evaluation of increasingly flexible reformulation capability.

\subsubsection{GSR-Dialect:}

We further expand \textbf{Level 2} with GSR-Dialect, a benchmark for transforming regional Chinese dialects into standard written Mandarin. This task introduces a broader form of semantic reformulation: models must normalize dialect‑specific pronunciations, lexical choices, and syntactic patterns while preserving the full meaning of the source utterance. The output may diverge substantially from the spoken surface, but entities, events, numerical values, temporal relations, negations, and other critical information must remain grounded in the source speech.

GSR-Dialect covers four representative dialect groups: Cantonese (Yue); Shanghainese (Wu); Sichuanese (Southwestern Mandarin); Henan dialect (Central Plains Mandarin).
The dataset combines publicly available dialect resources~\cite{dai2026wenetspeech, li2026wenetspeech} and proprietary business-oriented evaluation scenarios.
For proprietary content, personally identifiable information—names, locations, and sensitive business entities—is anonymized or replaced with synthetic alternatives before scripts are rendered into speech via dialect‑aware TTS. Native speakers then assess each synthesized sample on a five-point scale along two dimensions: semantic naturalness, which measures whether the utterance follows natural dialectal expressions and rhythm, and accent authenticity, which evaluates whether regional pronunciation characteristics are faithfully reproduced. Only samples receiving scores above 4.0 on both dimensions are retained. Since many dialects lack standardized written forms, native bilingual annotators provide standard Mandarin translations as reference outputs.

\subsubsection{GSR-Summary:}

\textbf{Level 3} encompasses a broad range of speech-grounded structuring tasks, such as summarization, intent extraction, and structured information generation. Since structured Level-3 capabilities are not consistently supported across existing LLM-based ASR systems, We instantiate this level with GSR-Summary, a focused benchmark for speech-grounded summarization, leaving broader structured-generation tasks for future extensions of GSR-Bench.

GSR-Summary is derived from ContextASR-Bench~\cite{wang2025contextasr}, whose long-form, multi-domain speech samples contain dense distributions of named entities and technical terms. These characteristics make its longer samples naturally suitable for evaluating whether a model can identify salient information from entity-intensive speech rather than merely reproduce the complete transcript.

However, raw benchmark is unsuitable for direct summarization evaluation due to inconsistent narratives, implausible entities, and unsupported facts. We apply a quality filter (logical consistency, factual plausibility, coherence, and suitability), retaining ~700 cases. The benchmark then tests whether models generate concise summaries that preserve key semantic units—entities, events, times, values, causality, negations, uncertainty—without introducing unsupported content.

\subsection{Grounding Reliability Evaluation}

Greater reformulation flexibility improves usability but also raises the risk of generation driven by linguistic priors rather than acoustic evidence. We therefore construct a dedicated stress test suite to evaluate grounding reliability independently from reformulation capability—measuring whether models correctly handle absent or degraded speech evidence.

\subsubsection{Acoustic Hallucination Stress Test}

The suite covers four scenarios. The \textit{environmental noise} subset contains real‑world backgrounds (customer‑service and monitoring scenarios), testing robustness under partial signal corruption. The \textit{silence} subset contains no valid speech; the expected output is empty, and any lexical output counts as hallucination. The \textit{music} subset provides structured non‑speech (background music), testing whether models misinterpret non‑linguistic signals as speech. The \textit{weak‑speech} subset, derived from AISHELL‑1 by attenuating amplitude to 3\% and adding mild noise, is deliberately constructed such that the content is practically unintelligible to both humans and ASR systems; any generated lexical content is therefore treated as acoustic hallucination, unlike conventional robustness benchmarks that focus on recognition accuracy under degradation.

\section{Results and Analysis}

We organize the experiments according to the five evaluation tracks of GSR-Bench:
(1) GSR-PACE (Level~0),
(2) GSR-Refine (Levels~1--2),
(3) GSR-Dialect (Leve~2),
(4) GSR-Summary (Level~3),
and (5) grounding reliability under acoustic stress.

We compare representative systems with different modeling paradigms, including conventional ASR, LLM-ASR models, ASR+LLM cascades, and general audio foundation models. 
To demonstrate the feasibility of unified grounded generation, we further develop \textbf{UniGSR}, an instruction-following Speech LLM based on a TeleASR speech encoder~\cite{chen2024telespeechpt} and Qwen3-4B~\cite{qwen3}.
Through instruction conditioning, UniGSR performs different speech understanding tasks within a single end-to-end architecture, including faithful transcription, refinement, dialect translation, and speech summarization.

Conventional ASR and LLM-ASR models mainly focus on Level~0 transcription and lack native reformulation ability.
ASR+LLM cascades extend these systems with text-based generation, but suffer from information bottlenecks because the LLM cannot directly access acoustic information.
General audio foundation models provide broader speech understanding capabilities and naturally support higher-level structuring tasks, but are not specifically optimized for grounded speech recognition.
In contrast, UniGSR is designed for grounded reformulation, enabling diverse textual representations while maintaining direct interaction with the original speech signal.

\begin{table*}[t]
\centering
\small
\caption{Multi-dimensional LLM-as-Judge evaluation on GSR-Refine. LLM-Scores range from 1 to 5; higher is better.}
\label{tab:meeting}
\begin{tabular}{lcccccccc}
\toprule
\multirow{2}{*}{\textbf{System}} & \multicolumn{3}{c}{\textbf{GPT-4o}} & \multicolumn{3}{c}{\textbf{DeepSeek-v4}} & \multirow{2}{*}{\textbf{LLM-Score $\uparrow$}} & \multirow{2}{*}{\textbf{CER $\downarrow$}}\\
\cmidrule(lr){2-4} \cmidrule(lr){5-7}
 & Fidelity & Fluency & Readability & Fidelity & Fluency & Readability & \\
\midrule
TeleASR~\cite{chen2024telespeechpt}    & 4.86  & 4.07  & 4.02  & \textbf{4.77} & 3.58 & 3.66 & 4.16 & 3.46\\
UniGSR-ASR          & 4.84   & 4.11 & 4.05  & 4.78 & 3.67 & 3.74 & 4.20 & 3.96\\
FireRed-LLM~\cite{xu2025fireredasr}  & 4.76 & 3.87 & 3.80 & 4.50 & 4.06 & 3.94 & 4.16 & 4.45\\
FireRed-AED~\cite{xu2025fireredasr}  & 4.73 & 3.86 & 3.78 & 4.47 & 4.05 & 3.91 & 4.13 & 6.06\\
FunASR-Nano~\cite{an2025funasrtechnicalreport}  & 4.65  & 3.85  & 3.74  & 4.62 & 3.71 & 3.69 & 4.04 & 6.10 \\
Qwen-ASR-1.7B~\cite{shi2026qwen3}  & 4.74  & 4.04  & 3.95 & 4.62 & 3.49 & 3.54 & 4.06 & 5.87\\
Qwen-ASR-0.6B~\cite{shi2026qwen3} & 4.60   & 3.98  & 3.83  & 4.23 & 4.10 & 3.81 & 4.09 & 6.79\\
\textbf{StepAudio-R1~\cite{tian2025step}} & 3.68  & 4.45  & 3.67  & 3.65  & 4.62  & 3.70  & 3.96 & 50.76 \\
\textbf{Gemini3.1 Pro~\cite{}} & 4.14  & 4.54  & 4.09  & 3.86  & 4.70  & 3.90  & 4.21 & 29.92\\
\textbf{UniGSR-Post}  & 4.45 & \textbf{4.95} & 4.45 & 4.29 & \textbf{4.74} & \textbf{4.23} & 4.52 & 54.44\\
\textbf{TeleASR $\rightarrow$ LLM (Cascade) }    & 4.80   & 4.84   & 4.71  & 4.65 & 4.13 & 4.08 & \textbf{4.54} & 16.28 \\
\bottomrule
\end{tabular}
\end{table*}

\subsection{L0: GSR-PACE}
\label{sec:results:context}

Table~\ref{tab:context} reports the performance of three representative systems under six context configurations. Since the target task remains faithful transcription, all improvements should arise solely from better utilization of contextual information rather than semantic reformulation. Therefore, a reduction in CER directly reflects successful context grounding.

\begin{table}[t]
\centering
\small
\caption{Context-conditioned transcription results (CER~\%, lower is better) under different context configurations. Bold marks the best result per system. Values below the No-Ctx baseline indicate successful context utilization; values above indicate degradation.}
\label{tab:context}
\begin{tabular}{lccc}
\toprule
\textbf{Context} & \textbf{FunASR} & \textbf{Qwen3-ASR-1.7B} & \textbf{UniGSR} \\
\midrule
No Context & 6.35 & 8.73 & 8.52 \\
Hotwords   & 4.30 & 7.97 & 4.79 \\
Domain     & 6.20 & 8.86 & 9.65 \\
Long Context & 5.87 & 6.26 & 5.73 \\
Dialogue   & 3.76 & 6.31 & 6.01 \\
User Profile & 6.18 & 8.85 & 17.00 \\
\bottomrule
\end{tabular}
\end{table}

As shown in Table~\ref{tab:context}, all three systems benefit from appropriate contextual information, demonstrating that external context can effectively resolve acoustically ambiguous expressions. 
Among different context types, hotwords, long context, and dialogue history provide the most consistent improvements, as they directly constrain entity recognition and discourse-level understanding.

UniGSR achieves the largest improvement under hotword and long-context conditions, indicating that instruction-conditioned speech generation can effectively incorporate external knowledge for context-aware recognition. 
However, inappropriate or overly specific context (e.g., user profile) may introduce conflicts with acoustic evidence and degrade performance, highlighting the importance of maintaining acoustic grounding when leveraging contextual information.

\subsection{L1\&2: GSR-Refine}
\label{sec:results:meeting}

Table~\ref{tab:meeting} jointly reports conventional CER and multi-dimensional LLM-as-a-Judge scores on the refined meeting transcription benchmark. While CER evaluates literal lexical fidelity, the LLM judges assess semantic preservation, expression fluency, and overall readability after reformulation. The discrepancy between the two metrics clearly illustrates the Metric Paradox: systems with nearly identical or even worse CER may produce substantially more readable transcripts.

As shown in Table~\ref{tab:meeting}, conventional CER and LLM-based evaluation exhibit substantially different trends. While CER measures literal lexical consistency with the reference, refined transcription explicitly aims to improve readability by inserting punctuation, performing inverse text normalization (ITN), and removing disfluencies. These operations inevitably modify the surface form of the transcript, leading to higher CER despite producing text that is more natural and easier to read.

The LLM-as-a-Judge evaluation addresses this limitation by assessing semantic fidelity, expression fluency, and readability from a generation perspective. For example, both UniGSR-Post and the TeleASR$\rightarrow$LLM cascade achieve the highest LLM scores despite exhibiting much higher CER than conventional ASR systems, demonstrating that lexical overlap alone is insufficient for evaluating grounded reformulation. To further validate the reliability of the LLM-based evaluation, we additionally conduct a human preference study, whose detailed results are provided in the supplementary material.

\subsection{L2: Cross-Dialect Semantic Reformulation}
\label{sec:results:dialect}

Table~\ref{tab:dialect} reports the performance of the cascade pipeline and UniGSR on cross-dialect semantic reformulation across four representative Chinese dialects. We evaluate both conventional machine translation metrics (BLEU, COMET, BERTScore, and Sentence-BERT) and semantic evaluation using two independent LLM judges (GPT-4o and DeepSeek-v4-flash).
\begin{table*}[t]
\centering
\small
\caption{Evaluation of Level-2 cross-dialect semantic reformulation.
BLEU, COMET, BERTScore, and Sentence-BERT measure lexical or embedding similarity against a single reference, whereas GPT-4o and DeepSeek-v4 evaluate semantic quality and translation naturalness (1--5, higher is better).}
\label{tab:dialect}
\begin{tabular}{llcccccc}
\toprule
\textbf{System} & \textbf{Dialect} & \textbf{BLEU} & \textbf{COMET} & \textbf{BERTScore} & \textbf{Sent-BERT} & \textbf{LLM (4o)} & \textbf{LLM (v4)} \\
\midrule
\multirow{4}{*}{TeleASR $\rightarrow$ LLM (Cascade)}
  & Yue               & 0.330 & 0.873 & 0.884 & 0.892 & 3.32 & 3.19 \\
  & Wu                & 0.532 & 0.914 & 0.934 & 0.932 & 3.89 & 3.62 \\
  & SW Mandarin       & 0.526 & 0.833  & 0.901   & 0.882   & 3.2 & 3.25  \\
  & Henan             & 0.307  & 0.809  & 0.860   & 0.859 & 2.98 & 3.08  \\
\midrule
\multirow{4}{*}{UniGSR}
  & Yue               & 0.297 & 0.849 & 0.847 & 0.872 & 3.61 & 3.36 \\
  & Wu                & 0.366 & 0.895 & 0.905 & 0.911 & 4.25 & 4.03 \\
  & SW Mandarin       & 0.619  & 0.872 & 0.928  & 0.926 & 3.78 & 3.42 \\
  & Henan             & 0.304 & 0.814 & 0.866 & 0.878  & 3.69 & 3.39 \\
\bottomrule
\end{tabular}
\end{table*}

Table~\ref{tab:dialect} shows that UniGSR achieves competitive performance against the cascade ASR+LLM system on cross-dialect semantic reformulation. 
While the cascade pipeline obtains slightly better scores on some reference-based metrics, these metrics rely on a single translation reference and may not fully reflect semantic equivalence. 
The LLM-based evaluation consistently favors UniGSR, especially on Wu and Henan dialects, indicating better semantic preservation and more natural reformulation.

This result demonstrates the advantage of end-to-end grounded generation for dialect scenarios. 
Unlike cascade systems that first compress dialect speech into an intermediate transcript and then perform text-only translation, UniGSR directly conditions generation on acoustic representations, allowing the model to better exploit dialect-specific pronunciation patterns and speech information.

\subsection{L3: Speech-grounded Structuring}
Table~\ref{tab:summary} reports the LLM-as-Judge evaluation results for Level~3 speech-grounded summarization. 
Unlike lower reformulation levels, Level~3 requires semantic abstraction and information compression while maintaining grounding with the original speech.

\begin{table}[t]
\centering
\small
\caption{LLM-as-Judge evaluation for Level-3 speech-grounded summarization. Five quality dimensions are scored from 1 to 5, where higher scores indicate better summary quality.}
\label{tab:summary}
\begin{tabular}{lccc}
\toprule
\textbf{Metric} & \textbf{Gemini 3.1 Pro} & \textbf{StepAudio-R1} & \textbf{UniGSR} \\
\midrule
Faithfulness      & 4.41 & 3.79 & 3.14 \\
Coverage          & 4.85 & 4.58 & 4.21 \\
Key Information   & 4.66 & 4.18 & 3.53 \\
Conciseness       & \textbf{4.95} & 4.84 & 4.74 \\
Fluency           & \textbf{4.97} & 4.87 & 4.60 \\
\bottomrule
\end{tabular}
\end{table}

General-purpose audio-language models, including Gemini~3.1 Pro and StepAudio-R1, achieve stronger performance, demonstrating their advantage in long-context speech understanding and high-level reasoning. 
However, their goal differs from GSR: general audio models aim to answer diverse questions about speech, whereas GSR aims to represent speech in useful textual forms without introducing information beyond the acoustic evidence.

UniGSR demonstrates the feasibility of extending a unified speech generation model toward Level~3 capability through instruction conditioning. Although further improvement is needed for complex summarization, these results show that a single GSR model can support tasks ranging from transcription to higher-level speech structuring.
General audio-language models aim to answer anything about speech, whereas GSR models aim to represent speech in every useful textual form without saying anything beyond it. 

\subsection{Grounding Reliability and Hallucination Analysis}
\label{sec:results:hallucination}

Table~\ref{tab:hallucination} reports grounding reliability under four acoustically challenging conditions, including environmental noise, silence, light music, and severely degraded speech. Following Section~4, the metric records the total number of generated characters under each condition. For the first three subsets, the expected output is empty, and any generated text is regarded as acoustic hallucination. For the weak-speech subset, the metric reflects the model's ability to preserve valid speech content without over-generating unsupported text.

\begin{table}[t]
\centering
\small
\caption{Grounding reliability under acoustic stress. The evaluation suite contains 520 audio segments spanning environmental noise, silence, music, and unintelligible weak speech. Any non-empty output is counted as an acoustic hallucination. Lower is better.}
\label{tab:hallucination}
\begin{tabular}{lccc}
\toprule
\textbf{System}
&
\textbf{HTR (\%)$\downarrow$}
&
\textbf{AHL$\downarrow$}
&
\textbf{Total Chars$\downarrow$}
\\
\midrule
UniGSR           & 15.0 & 17.28 & 1348 \\
FunASR-Nano      & 6.9 & 54.08 & 1947 \\
Qwen-ASR-0.6B    & 14.4 & 11.33 & 850 \\
Qwen-ASR-1.7B    & 9.8 & 10.31 & 526 \\
FireRed          & 16.0 & 14.16 & 1175 \\
Gemini3.1 Pro    & 10.8 & 12.57 & 704 \\
StepAudio-R1     & \textbf{5.4} & \textbf{8.79} & \textbf{246} \\
\bottomrule
\end{tabular}
\end{table}

As shown in Table~\ref{tab:hallucination}, hallucination behaviors vary substantially across different speech generation paradigms. 
StepAudio-R1 achieves the lowest hallucination rate and shortest hallucinated outputs, followed by Qwen-ASR-1.7B, indicating strong suppression of unsupported generation under acoustic stress. 
In contrast, models with stronger generative capabilities may produce more hallucinated content when the acoustic evidence is insufficient.

UniGSR shows a moderate hallucination level while supporting broader reformulation capabilities beyond conventional ASR, including semantic rewriting and speech structuring. 
This result suggests that maintaining acoustic grounding is challenging when extending speech models from recognition toward open-ended generation. 
Therefore, GSR requires not only generating useful textual representations from speech, but also explicitly constraining generation to avoid introducing information unsupported by the acoustic input.

\section{Conclusion}

In this paper, we introduced \emph{Generative Speech Recognition (GSR)}, a new speech recognition paradigm that extends conventional transcription to grounded reformulation. We formalized GSR through the concepts of the Grounded Reformulation Space and a four-level Reformulation Hierarchy, providing a unified framework that encompasses faithful transcription, canonical normalization, semantic reformulation, and speech-grounded structuring.

To support systematic evaluation, we presented \emph{GSR-Bench}, the first benchmark designed to assess grounded generative speech recognition across multiple reformulation levels and grounding robustness. Experimental results on representative ASR systems, cascade ASR+LLM pipelines, Speech LLMs, and general audio foundation models demonstrate that existing systems exhibit substantially different capability boundaries under the proposed framework, highlighting the need for evaluation beyond conventional transcription metrics. We hope that GSR and GSR-Bench will provide a common foundation for future research on instruction-following and grounded speech generation.

\bibliography{aaai2027}


\end{document}